\documentclass{article}
\usepackage{xcolor}
\usepackage[preprint]{corl_2026}
\usepackage{amsmath,amssymb}
\usepackage{array}
\usepackage{booktabs}
\makeatletter\newcommand{\captionof}[1]{\def\@captype{#1}\caption}\makeatother
\usepackage{enumitem}
\usepackage{float}
\usepackage{graphicx}
\usepackage{xcolor}
\usepackage{colortbl}
\definecolor{brandpurple}{HTML}{7B236D}
\definecolor{hlrow}{HTML}{F3E8F1}
\newcommand{\yes}{\ensuremath{\checkmark}}
\newcommand{\no}{\ensuremath{\times}}
\newcommand{\planned}{\ensuremath{\circ}}
\usepackage{xurl}
\usepackage{microtype}

\makeatletter
\newcommand{\code}[1]{%
  \begingroup\ttfamily
  \renewcommand{\_}{\char`\_\allowbreak}%
  \hu@slash#1/\hu@stop\hu@nil
  \endgroup}
\def\hu@slash#1/#2\hu@nil{%
  #1%
  \def\hu@arg{#2}\def\hu@end{\hu@stop}%
  \ifx\hu@arg\hu@end\else/\allowbreak\hu@slash#2\hu@nil\fi}
\makeatother
\newcommand{\systemname}{HandUMI}

\newcommand{\statTowelNumEpisodes}{562}
\newcommand{\statTowelNumFrames}{201{,}397}
\newcommand{\statTowelNumRowsKept}{198{,}025}

\newcommand{\statTowelTotalMinutes}{111.9}

\newcommand{\statTowelSyncLwristMedMs}{8.5}
\newcommand{\statTowelSyncLwristPNinetyFiveMs}{16.4}

\newcommand{\statTowelSyncLwristOverLimitCount}{3}
\newcommand{\statTowelSyncRwristMedMs}{8.6}
\newcommand{\statTowelSyncRwristPNinetyFiveMs}{16.3}

\newcommand{\statTowelSyncRwristOverLimitCount}{2}
\newcommand{\statTowelSyncWorkspaceMedMs}{8.9}
\newcommand{\statTowelSyncWorkspacePNinetyFiveMs}{16.2}

\newcommand{\statTowelSyncWorkspaceOverLimitCount}{2}
\newcommand{\statTowelSyncEncoderMedMs}{2.5}
\newcommand{\statTowelSyncEncoderPNinetyFiveMs}{4.8}

\newcommand{\statTowelSyncEncoderOverLimitCount}{4}
\newcommand{\statTowelSyncTrackingMedMs}{40.3}
\newcommand{\statTowelSyncTrackingPNinetyFiveMs}{40.4}

\newcommand{\statTowelSyncTrackingOverLimitCount}{0}

\newcommand{\statTowelTrackedPct}{100.00}

\title{\systemname{}: A Modular Hand-Worn Parallel-Jaw Interface for Robot-Free Bimanual Data Capture}

\author{\texorpdfstring{%
    Brik Meza$^{1,2,3}$ \quad Leonardo Perez$^{1,4,5}$ \quad Alvaro Mendoza-Li$^{1,6}$\\
    \bfseries Aniela Ubillus$^{1,4,5}$ \quad Mitshell Ramos$^{4,5}$ \quad Raul Bastidas$^{5,7}$\\
    \bfseries Mauricio Olivera$^{4,5}$ \quad Bryan Bastidas$^{4,5}$\\[3pt]
    {\normalfont\small $^{1}$Mu Robotics \quad
      $^{2}$McGill University \quad
    $^{3}$Mila - Quebec AI Institute}\\
    {\normalfont\small $^{4}$Pontificia Universidad Cat\'olica del Per\'u (PUCP) \quad
    $^{5}$NONHUMAN}\\
    {\normalfont\small $^{6}$Technical University of Munich (TUM) \quad
    $^{7}$Universidad Peruana de Ciencias Aplicadas (UPC)}%
}{Brik Meza, Leonardo Perez, Alvaro Mendoza-Li, Aniela Ubillus, Mitshell Ramos, Raul Bastidas, Mauricio Olivera, Bryan Bastidas}}

\makeatletter
\hypersetup{%
  pdftitle={\@title},
  pdfauthor={\@author},
  pdfsubject={The UMI Arena, CoRL 2026 Workshop, Track 3 (non-archival)}%
}
\makeatother

\begin{document}
\maketitle

\begin{abstract}
\systemname{} captures robot-free bimanual demonstrations with a hand-worn
parallel-jaw interface. Swappable fingertips match the robot's installed
gripper; Quest/PICO controllers track hands and encoders measure aperture.
Robot-agnostic recordings are retargeted by
inverse kinematics to seven simulated bimanual arms (median position error
$\le$1.2\,mm; $\le$4.7\% of episodes rejected). On bimanual towel folding,
\statTowelNumEpisodes{} robot-free demonstrations train policies on real PiPER
arms: ACT succeeds in 12/20 attempts and fine-tuned $\pi_{0.5}$ in 7/20;
ACT on the first 101 and base $\pi_{0.5}$ each succeed in 0/20. Hardware and
software are open source (Apache-2.0), and a pair costs USD~557 with Quest~3S,
excluding the optional ZED~Mini and host computer.
Project website: \url{https://murobotics.ai/research/handumi}.
\end{abstract}


\section{Introduction}
\label{sec:intro}

Robot teleoperation with leader arms, replicas or VR
controllers~\citep{khazatsky2024droid,zhao2023act,fu2024mobilealoha,wu2024gello}
requires robot access for each demonstration.

Robot-free interfaces such as UMI~\citep{chi2024umi} decouple capture from arm
kinematics, but contact-geometry or opening-motion mismatch may hinder transfer.
YUBI~\citep{ohkawa2026yubi}, the closest, also tracks Quest controllers, measures
aperture with an encoder and is open. \systemname{} differs in three ways: the
robot keeps its gripper instead of being refitted with YUBI's; jaws translate,
whereas YUBI records a jaw angle and models revolute
fingers~\citep{ohkawa2026yubi,yubi2026urdf}; and tips copy installed grippers,
whereas its paper describes one finger design. We do not compare them directly. KIWI~\citep{yang2026kiwi} offers interchangeable
tool modules but estimates pose by offline SLAM and jaw width from video, with its
software release only announced. FastUMI~\citep{zhaxizhuoma2024fastumi} adapts
fingertips, and ActiveUMI~\citep{zeng2025activeumi} copies existing grippers in
controller-held interfaces. Table~\ref{tab:compare} summarizes these choices.

\begin{figure}[t]
  \centering
  \includegraphics[width=\linewidth]{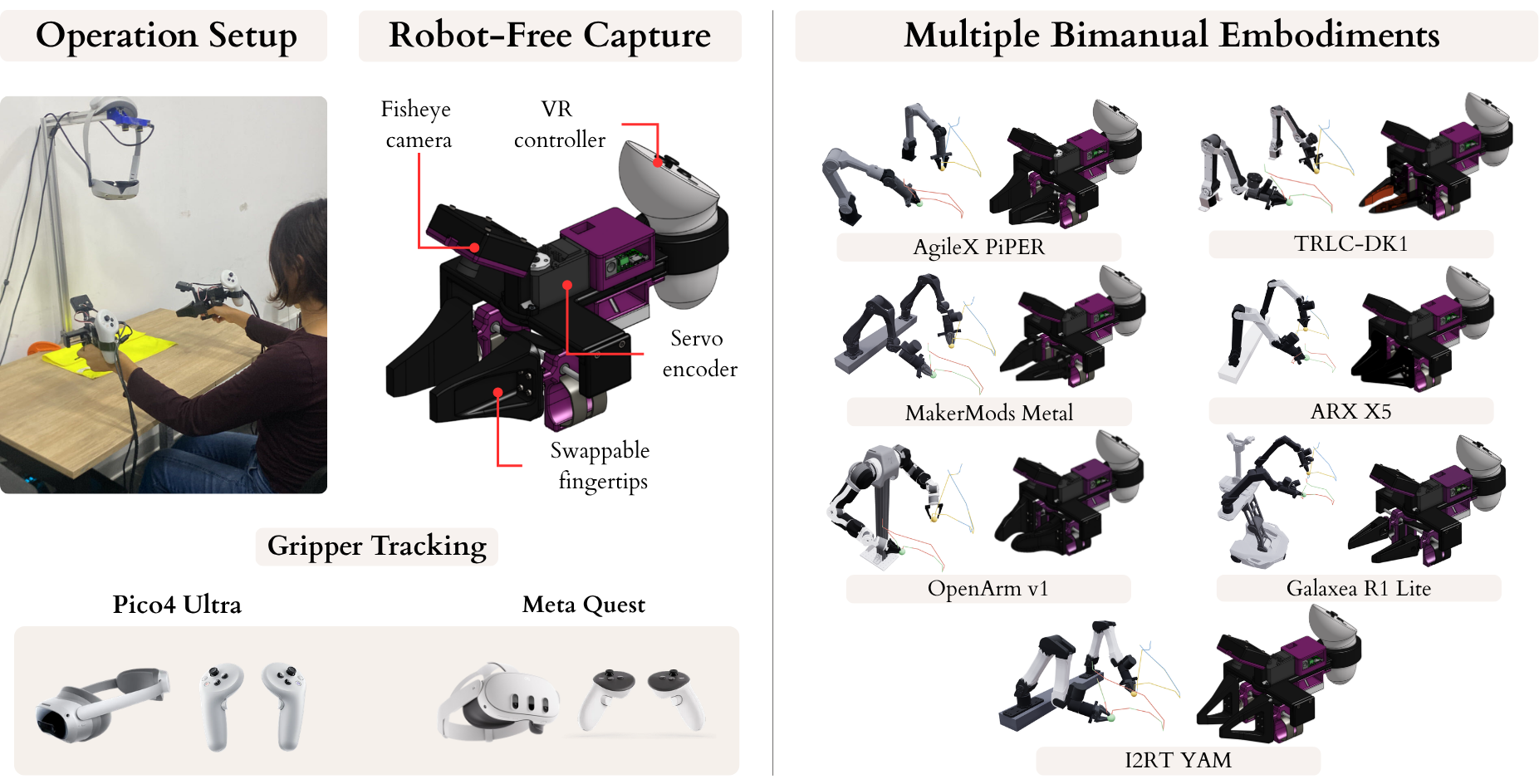}
  \caption{\systemname{}. Left: robot-free capture with the hand-worn device
  (fisheye camera, servo encoder, swappable fingertips) tracked by PICO~4 Ultra or
  Meta Quest controllers. Right: recorded trajectories retargeted in simulation to
  seven bimanual arms.}
  \label{fig:device}
\end{figure}

\systemname{} (Fig.~\ref{fig:device}) contributes:
\begin{itemize}[leftmargin=*,itemsep=1pt,topsep=3pt,parsep=0pt]
  \item An open, low-cost hand-worn interface with translating parallel jaws,
        robot-matched swappable fingertips, live controller tracking and
        encoder-measured aperture; the robot retains its installed gripper.
  \item A pipeline preserving robot-agnostic demonstrations and calibration
        provenance, with inverse kinematics (IK) and screening for seven
        bimanual arms.
  \item Real-robot evidence on bimanual towel folding with PiPER arms: ACT~\citep{zhao2023act}
        and fine-tuned $\pi_{0.5}$~\citep{black2025pi05}, trained only on
        retargeted \systemname{} data, fold the towel in unassisted rollouts.
\end{itemize}

\begin{table}[H]
  \centering
  \footnotesize
  \setlength{\tabcolsep}{2.1pt}
  \caption{Robot-free interfaces. Enc.: encoder-measured gripper state; Open HW/SW: public files/code; Tips: swappable tips.
  \yes/\no: yes/no; \planned: announced; --: not established; n/a: not
  applicable. USD: named parts for a pair with tracking, excluding host, tax
  and shipping; $>$: only the headset is priced.}
  \label{tab:compare}
  \begin{tabular}{@{}llccccccr@{}}
    \toprule
    Device & Pose source & Enc. & Bimanual & Open HW & Open SW & Tips & Audio & USD \\
    \midrule
    UMI~\citep{chi2024umi} & mono-inertial SLAM & \no & \yes & \yes & \yes & -- & -- & 742$^a$ \\
    UMI-3D~\citep{wang2026umi3d} & LiDAR-inertial odom. & \no & \no & \yes & \yes & -- & -- & 1{,}400$^b$ \\
    FastUMI~\citep{zhaxizhuoma2024fastumi} & T265 VIO & \no & \yes & -- & \yes$^c$ & \yes & -- & $\approx$2{,}042$^d$ \\
    DexUMI~\citep{xu2025dexumi} & iPhone ARKit & \yes & --$^e$ & \yes & \yes & n/a & -- & $\approx$2{,}448$^f$ \\
    iPhUMI~\citep{patel2026iphumi} & iPhone ARKit & \no & \yes & \yes & \yes & -- & \yes$^g$ & $\approx$1{,}504$^h$ \\
    RDT2~\citep{liu2026rdt2} & Vive Tracker (IR) & -- & \yes & -- & \yes$^i$ & -- & -- & $\approx$1{,}750$^j$ \\
    HiFi-UMI~\citep{wei2026hifiumi} & head SLAM + tags & \yes & \yes & -- & -- & -- & -- & -- \\
    YUBI~\citep{ohkawa2026yubi} & Quest ctrl. & \yes & \yes & \yes & \yes & -- & -- & $<$750$^k$ \\
    ActiveUMI~\citep{zeng2025activeumi} & Quest ctrl. & -- & \yes & -- & -- & -- & -- & $>$350$^l$ \\
    EgoMI~\citep{yu2025egomi} & Quest 3S ctrl. & -- & \yes & \planned & \planned & -- & -- & $\approx$12{,}500$^m$ \\
    XRZero-G0~\citep{wang2026xrzero} & PICO~4 ctrl. & -- & \yes & -- & --$^n$ & -- & -- & $>$463$^o$ \\
    KIWI~\citep{yang2026kiwi} & 360$^\circ$ SLAM + IMU & \no & \yes & \yes & \planned & \yes$^p$ & \yes & $\approx$1{,}100$^q$ \\
    \rowcolor{hlrow}
    \textbf{\systemname{}} & Quest / PICO ctrl. & \yes & \yes & \yes & \yes & \yes & \yes$^r$ & \textbf{557}$^s$ \\
    \bottomrule
  \end{tabular}\\[2pt]
  \parbox{\linewidth}{\footnotesize\raggedright
  $^a$2$\times$(gripper USD~73 + GoPro kit USD~298), historical.
  $^b$2$\times$(gripper USD~50 + LiDAR/camera USD~650).
  $^c$Capture code and dataset. $^d$2$\times$(GoPro~9 + T265) + HD60~X.
  $^e$Unimanual evaluation.
  $^f$2$\times$(iPhone~15 Pro Max + OAK-1~W + electronics); one hand published.
  $^g$Contact microphone. $^h$2$\times$(iPhone~15 Pro + UMI gripper).
  $^i$Models and training/inference code.
  $^j$4 Vive Trackers + 2 Hikrobot cameras.
  $^k$2$\times$``under USD~200'' + Quest~3S (USD~349.99). $^l$Quest~3S only.
  $^m$2$\times$Robotiq 2F-85 + Quest~3S + ZED~2i. $^n$Dataset only.
  $^o$PICO~4 only. $^p$Tool modules. $^q$Two Insta360 X5; grippers unpriced.
  $^r$Optional PICO audio, unused here.
  $^s$USD~207.11 parts + Quest~3S; optional ZED~Mini excluded.}
\end{table}

\section{\systemname{}}
\label{sec:system}

\paragraph{Hand-worn hardware.} Modular mounts allow swapping fingertips,
controller types and cameras on a common wearable body. Printed finger cradles drive parallel jaws
through a crank (Fig.~\ref{fig:device}). A Feetech STS3215 12-bit encoder
measures jaw width via two-point calibration; an IMX335 fisheye records 1280$\times$960 at
30\,fps; an optional ZED~Mini adds a workspace view. Fingertips match
AgileX PiPER~\citep{agilex2024piper}, ARX~X5~\citep{arx2025x5}, the
Dream Gripper~\citep{dreammachines2026dreamgripper} for TRLC-DK1, Trossen WidowX~AI~\citep{trossen2026widowxai}
and the UMI soft finger~\citep{chi2024umi}; the robot retains its gripper.
Parts cost USD~207.11 per pair, or USD~557 with shared
Quest~3S~\citep{meta2026questpricing}, excluding the optional ZED~Mini,
host computer, tax and shipping. All experiments used PICO~4 Ultra.

\paragraph{Tracking, synchronization and calibration.} Instead of wrist-camera SLAM, a
swappable provider supplies timestamped 6-DoF controller poses: Quest~3/3S through
a new TCP app implementing the protocol of YUBI's Apache-2.0
software~\citep{ohkawa2026yubi}, informed by analysis of its client APK, or PICO~4 Ultra
through XRoboToolkit~\citep{zhao2026xrobotoolkit}. A ChArUco
board~\citep{opencv2026charuco,garridojurado2014aruco} defines the table frame. The
workstation's monotonic clock stamps every source; rows target 40\,ms behind real
time and take the nearest camera (30\,fps) and encoder (100\,Hz) samples with
freshness flags (60\,ms skew limit); sensor loss over 1\,s discards the episode.
Voice commands~\citep{vosk} or double-squeezes control recording.
Gated calibration covers gripper range, intrinsics
($\le$0.8\,px mean reprojection error),
camera mount and table frame ($\le$8\,mm RMS), and per-tip tool center point
(TCP) pivot fits ($<$5\,mm RMS, $<$10\,mm maximum sample residual).
A separate guided touch protocol checks both tips at one reference point
(15\,mm separation limit) and on the table ($\pm$15\,mm height), using
20 samples per position by default; no result can be tied to the
affected sessions. SHA-256 fingerprints identify configurations, not accuracy.

\paragraph{Source-preserving retargeting.} LeRobot~v3
recordings~\citep{cadene2024lerobot,cadene2026lerobot} store two controller poses
and jaw widths independently of the robot. Profiles supply kinematics,
limits, TCPs and a linear gripper map (0--66\,mm onto PiPER's
stroke; speed and force unmatched). PyRoki IK~\citep{kim2025pyroki} generates
joint data for PiPER, OpenArm~v1~\citep{enactic2025openarm},
TRLC-DK1~\citep{trlc2025dk1}, I2RT YAM~\citep{i2rt2025yam}, MakerMods Metal
Arm~\citep{makermods2026metal}, Galaxea
R1~Lite~\citep{galaxea2026r1lite,jiang2025galaxea} and ARX~X5~\citep{arx2025x5}.
Per-robot screening checks IK error, self-collision
and base-joint swing. PiPER and OpenArm replay
shares mapping, IK and smoothing with MuJoCo~\citep{todorov2012mujoco}.

\section{Experiments}
\label{sec:eval}

We evaluate \systemname{} on bimanual PiPER towel folding: the arms grasp two
corners of a flat towel and fold it in half.

\subsection{Capture quality}

\paragraph{Dataset.} Without a robot, three operators (authors; one with prior
device experience) used PICO/PiPER-tip \systemname{} to record the cohort over
four days (about 5\,h in total). Besides 3 unfolded towels and 1 tracking
freeze, we dropped 79 of 641 episodes (12\%) whose PiPER retargeting swings
a base joint past 60$^\circ$, biasing it toward PiPER's workspace; the other
arms accept 64--79 of these. The \statTowelNumEpisodes{} kept
episodes (\statTowelNumFrames{} frames, \statTowelTotalMinutes{}\,min) pass the
offline episode-quality gate: duration, finite values, tracking, sensor health,
synchronization, pose jumps, freezes and aperture checks. Tracking, health and
sync each allow up to 1\% bad rows; health and sync exclude the first 0.2\,s.
A single row above 60\,ms does not reject an episode. After faulty camera
recalibration, 40 episodes received a post-hoc +1.9\,cm height correction
estimated from grasp contacts; 31 remain in the cohort. Fit residuals and
offline checks do not independently verify physical table height; the
reason the gates missed the fault is unverified.

\begin{table}[!htb]
  \footnotesize
  \setlength{\tabcolsep}{3pt}
  \begin{minipage}[t]{0.38\linewidth}
    \centering
    \caption{Selection skew over \statTowelNumRowsKept{} rows
    (warm-up excluded).}
    \label{tab:sync}
    \begin{tabular}{@{}lrrr@{}}
      \toprule
      Stream & Med. & p95 & $>$60\,ms \\
      & (ms) & (ms) & (rows) \\
      \midrule
      Left wrist cam. & \statTowelSyncLwristMedMs & \statTowelSyncLwristPNinetyFiveMs & \statTowelSyncLwristOverLimitCount \\
      Right wrist cam. & \statTowelSyncRwristMedMs & \statTowelSyncRwristPNinetyFiveMs & \statTowelSyncRwristOverLimitCount \\
      Top camera & \statTowelSyncWorkspaceMedMs & \statTowelSyncWorkspacePNinetyFiveMs & \statTowelSyncWorkspaceOverLimitCount \\
      Encoder (100\,Hz) & \statTowelSyncEncoderMedMs & \statTowelSyncEncoderPNinetyFiveMs & \statTowelSyncEncoderOverLimitCount \\
      PICO tracking & \statTowelSyncTrackingMedMs & \statTowelSyncTrackingPNinetyFiveMs & \statTowelSyncTrackingOverLimitCount \\
      \bottomrule
    \end{tabular}
  \end{minipage}\hfill
  \begin{minipage}[t]{0.60\linewidth}
    \centering
    \caption{Retargeting of 641 episodes (Metal: 79). Position error is the median.}
    \label{tab:retarget}
    \setlength{\tabcolsep}{3pt}
    \begin{tabular}{@{}lrrrr@{}}
      \toprule
      Robot & Position & Rejected & Self- & Joint \\
      & error (mm) & (\%) & collision (\%) & limit (\%) \\
      \midrule
      PiPER & 1.17 & 0.0 & 0.0 & 0.1 \\
      OpenArm & 0.14 & 4.7 & 0.0 & 27.5 \\
      TRLC-DK1 & 0.18 & 3.1 & 2.7 & 2.3 \\
      YAM & 0.19 & 0.8 & 0.0 & 4.7 \\
      Metal & 0.09 & 0.0 & 0.0 & 0.4 \\
      R1~Lite & 0.03 & 0.0 & 16.4 & 0.0 \\
      ARX X5 & 0.09 & 0.0 & 0.0 & 0.1 \\
      \bottomrule
    \end{tabular}
  \end{minipage}
\end{table}

\paragraph{Synchronization and calibration.} Table~\ref{tab:sync} measures
sampling-grid skew ($\approx$8\,ms median at 30\,fps), not end-to-end latency.
PICO leads $\approx$40\,ms on the host clock; both wrists are tracked on
\statTowelTrackedPct\% of rows. TCP fits (750 samples/side) pass both thresholds:
RMS 4.0/3.4\,mm; maximum sample-to-fitted-pivot Euclidean residuals
7.013/7.021\,mm (left/right). These are internal consistency, not absolute accuracy.
Table-frame RMS is 1.6--4.2\,mm and 0.3--0.7$^\circ$; two remounts differ
6--7\,mm in TCP height.

\paragraph{Cross-embodiment retargeting.} Table~\ref{tab:retarget} replays 562
kept and 79 dropped episodes on seven simulated arms (Metal: only the 79, after a
compute failure). PiPER uses the lab table; other placements are hyperparameters,
not held out (OpenArm, ARX X5: swept on 19, 10 evaluated episodes). Rejected: unreachable
start or p99 position error $>$3\,cm. Self-collision: hull-confirmed episodes.
Joint limit: frames at a limit. Up to 22.9\% of episodes exceed URDF velocity
limits (OpenArm). OpenArm is 7-DoF, others 6-DoF. PiPER-tip data test kinematics, not
contact; PiPER's rotation residual enters policy training, with unknown effect.

\subsection{Robot-free data for policies}

\begin{figure}[!htb]
  \centering
  \includegraphics{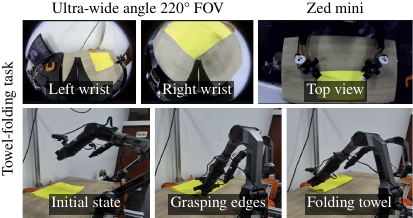}\hspace{0.2in}
  \includegraphics{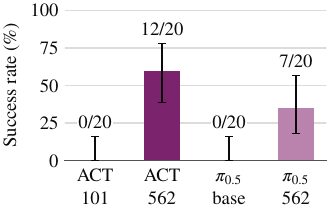}
  \caption{Towel folding: views, a rollout, and successes by training episodes (Wilson 95\% CI).}
  \label{fig:policy}
\end{figure}

\paragraph{Setup.} Policies use \systemname{} IK commands as 14-D PiPER
states/actions (radians, grippers in [0,\,1]) and two wrist views, excluding
the top view (Fig.~\ref{fig:policy}). PiPER retains the same IMX335 cameras
and gripper-relative positions. Real PiPER open-loop replay validated recorded
trajectories. ACT~\citep{zhao2023act} uses one recipe for 101 episodes
(19.3\,min) and the full cohort; $\pi_{0.5}$~\citep{black2025pi05} uses full-cohort
fine-tuning and a base-model reference. A 10\% episode holdout selects the
lowest-validation-loss checkpoint (ACT: batch 96, lr $10^{-5}$, chunk 100;
$\pi_{0.5}$: batch 16, lr $4{\times}10^{-5}$, chunk 50).

\paragraph{Protocol.} Each model runs 20 unassisted attempts at random towel
positions within the training area, unmatched across models. The operator
judges a final fold in half live, without numeric tolerance. Videos: project website;
data and checkpoints: \url{https://huggingface.co/murobotics}.

\paragraph{Results.} ACT on all 562 episodes succeeds in 12/20 attempts
(95\% Wilson CI 39--78\%) and fine-tuned $\pi_{0.5}$ in 7/20 (18--57\%;
Fig.~\ref{fig:policy}); at $n{=}20$ they are not statistically distinguishable
(Fisher exact $p{=}0.20$). ACT on the first 101 episodes and $\pi_{0.5}$ without
fine-tuning each succeed in 0/20 (0--16\%). The former is a scale probe,
confounded by added sessions/calibrations; the latter is a sanity check for
\systemname{} fine-tuning.

\section{Conclusion and Limitations}
\label{sec:conclusion}

\systemname{} abstracts arm kinematics while retaining the robot's gripper;
new grippers require new capture tips. Contact geometry remains
embodiment-dependent. Robot-free data enable real PiPER towel folding
(ACT 12/20, $\pi_{0.5}$ 7/20).

\paragraph{Limitations.} Cross-embodiment evidence is simulated IK; real tests cover one PiPER task,
setup and seed, without failure analysis; holdouts can share sessions. Capture is headset-bounded and parallel-jaw; tip-matching
gain, jaw accuracy, fingertip load and audio value are unmeasured.

\clearpage
\acknowledgments{We thank the NONHUMAN community for their collaboration on
data collection and device testing.}

\bibliography{references}

\end{document}